\pdfoutput=1

\documentclass[11pt]{article}

\usepackage{EACL2023}

\usepackage{times}
\usepackage{latexsym}

\usepackage{footnote}
\usepackage{array}
\usepackage{graphicx}
\usepackage{color, colortbl}
\usepackage{xstring}
\usepackage{comment}
\usepackage{adjustbox}
\usepackage{float}
\restylefloat{table}
\usepackage{array,booktabs,makecell}
\usepackage{multirow}
\usepackage{graphicx}
\usepackage{listings}
\usepackage{xcolor}

\usepackage{appendix}
\usepackage{times,latexsym}
\usepackage{url}
\usepackage[T1]{fontenc}
\usepackage{lipsum}  
\usepackage{graphicx}
\usepackage{xcolor} 
\usepackage{soul}
\usepackage{listings}
\usepackage[T1]{fontenc}

\usepackage[utf8]{inputenc}

\usepackage{microtype}

\colorlet{punct}{red!60!black}
\definecolor{background}{HTML}{EEEEEE}
\definecolor{delim}{RGB}{20,105,176}
\colorlet{numb}{magenta!60!black}

\lstdefinelanguage{json}{
    basicstyle=\normalfont\ttfamily,
    stepnumber=1,
    showstringspaces=false,
    breaklines=true,
    frame=lines,
    backgroundcolor=\color{white},
    literate=
     *{0}{{{\color{numb}0}}}{1}
      {1}{{{\color{numb}1}}}{1}
      {2}{{{\color{numb}2}}}{1}
      {3}{{{\color{numb}3}}}{1}
      {4}{{{\color{numb}4}}}{1}
      {5}{{{\color{numb}5}}}{1}
      {6}{{{\color{numb}6}}}{1}
      {7}{{{\color{numb}7}}}{1}
      {8}{{{\color{numb}8}}}{1}
      {9}{{{\color{numb}9}}}{1}
      {:}{{{\color{punct}{:}}}}{1}
      {,}{{{\color{punct}{,}}}}{1}
      {\{}{{{\color{delim}{\{}}}}{1}
      {\}}{{{\color{delim}{\}}}}}{1}
      {[}{{{\color{delim}{[}}}}{1}
      {]}{{{\color{delim}{]}}}}{1},
}

\title{Zero-Shot Slot and Intent Detection in Low-Resource Languages}

\author{\normalsize Sang Yun Kwon$^{1,\star}$ ~ Gagan Bhatia $^{1,\star}$ ~ El Moatez Billah Nagoudi$^{1}$ 
~\\ {\bf Alcides Alcoba Inciarte}$^{1}$ ~{\bf Muhammad Abdul-Mageed}$^{1,2}$\\
\normalsize $^{1}$Deep Learning \& Natural Language Processing Group,
  The University of British Columbia\\\normalsize  $^{2}$Department of Natural Language Processing \& Department of Machine Learning, MBZUAI\\ %
  \texttt{\normalsize \{skwon01@student.,gagan30@student.,moatez.nagoudi@,alcobaaj@mail.,muhammad.mageed@\}ubc.ca}}

\begin{document}
\maketitle
\begin{abstract}
Intent detection and slot filling are critical tasks in spoken and natural language understanding for task-oriented dialog systems. In this work we describe our participation in the slot and intent detection for low-resource language varieties  (SID4LR;~\citet{2023-findings-vardial}).  We investigate the slot and intent  detection 
 (SID) tasks using a wide range of models and settings. Given the recent success of multitask-prompted finetuning of large language models, we also test the generalization capability of the recent encoder-decoder model mT0~\cite{muennighoff2022crosslingual} on new tasks (i.e., SID)  in languages they have never intentionally seen. We show that our best model outperforms the baseline by a large margin (up to $+30$ F\textsubscript{1} points) in both SID tasks.

\end{abstract}
\section{Introduction}\label{sec:intro}


Digital conversational assistants have become increasingly pervasive. Examples of popular virtual assistants include Siri, Alexa, and Google. A crucial factor in the effectiveness of these systems is their capacity to understand user input and respond or act accordingly to fulfill particular requirements. Most of these applications are voice-based and hence need spoken language understanding (SLU). SLU typically starts with automatic speech recognition (ASR), taking the sound of spoken language and transcribing it into text. Then, it handles natural language understanding (NLU) tasks to extract semantic features from the text including question answering, dialogue management,  intent detection, and slot filling.

The intent detection task aims to recognize the speaker's desired outcome from a given utterance. And slot filling focuses on identifying the main arguments or the spans of words in the utterance that contain semantic information relevant to the intent. Table~\ref{tab:xsid_ex} shows four utterances in different languages: English, Swiss German (GSW), South Tyrolean (ST), and Neapolitan (NAP). The English example has the intent \textit{set\_alarm} and two individual spans \textit{Set an alarm} and \textit{6 am on Wed} are labeled with their slot tags \textit{location} and \textit{datetime}, respectively, using the Inside, Outside, Beginning (IOB) \cite{ramshaw-marcus-1995-text} tagging format.

\begin{table}
\centering
 \renewcommand{\arraystretch}{1.75}
\resizebox{1\columnwidth}{!}{%

\begin{tabular}{ll}
 \toprule
\textbf{Lang.} & \textbf{~~~~~~~~~~~~~~~~~~~~~~~~~~~~~~~~~~~~~~~~Annotation} \\  \toprule
\textbf{EN} & \colorbox{green!35}{Set an alarm} for\colorbox{orange!35}{6 am on Wed} \\
\textbf{GSW } & Du em \colorbox{green!35}{Mittwuch e Wecker} dry fürem sächsi em \colorbox{orange!35}{Morge}. \\ 
\textbf{ST} & Stell an \colorbox{green!35}{Wecker} firn \colorbox{orange!35}{Mittig af 6 in der friah} \\
\textbf{NAP} & Imposta 'na \colorbox{green!35}{sveglia} 'e \colorbox{orange!35}{6 'e matina 'e miercurì} \\ \toprule
\end{tabular}}
\caption{Examples of xSID annotations in our target languages from the validation set with intents (alarm / set\_alarm) and slots (\colorbox{green!35}{location}, \colorbox{orange!35}{datetime}). \textbf{EN}: English, \textbf{GSW}:Swiss German \textbf{ST}: South Tyrolean, \textbf{NAP}: Neapolitan} \label{tab:xsid_ex} 
\end{table}

In this work, we present our participation in the  slot and intent detection for low-resource language varieties (SID4LR;~\citet{2023-findings-vardial}) shared task. The shared task takes as its target  
three low resources languages-- Swiss German (GSW), South Tyrolean (ST), and Neapolitan (NAP).  
The main objective of the SID4LR shared task is to find the most effective approach for transferring knowledge to less commonly spoken languages that have limited resources and lack a standard writing system, in the zero-shot setting (i.e., without use of any training data). In the context of the shared task, we target the following four main research questions:

\begin{enumerate}

    \item[\textbf{Q1:}] Can successful models on English SID tasks be generalizable to new unseen languages (i.e., the zero-shot setting)?

      \item[\textbf{Q2:}] How do models trained  on a language from the given language family fare on a low-resource variety from the same family under the zero-shot setting (i.e., with no access to training data from these low-resource varieties). For example, in our case, we ask how do models trained on German perform on Swiss German or South Tyrolean, and how do models trained on Italian perform on Neapolitan. 

\item[\textbf{Q3:}] What impact does exploiting data augmentation techniques such as  paraphrasing and machine translation have on the SID tasks in the zero-shot context?

   \item[\textbf{Q4:}] Are the existing large multilingual models, trained using multitask-prompted fine-tuning, able to achieve zero-shot generalization to SID tasks in languages that they have never intentionally seen?


\end{enumerate}

The rest of this paper is organized as follows: Section~\ref{sec:RW} is a literature review on intent and slot  detection tasks. The shared task, the source data provided in SID4LR, and  the external parallel data we exploit to build our models  are described in Section ~\ref{sec:SID4LR}. In Section~\ref{sec:data}, we provide information about datasets, baselines, and data preprocessing. The baseline, and multilingual pre-trained language models we used are described in Section~\ref{sec:models}. We present our experimental settings and our training procedures in Section~\ref{sec:exp}. Section~\ref{sec:results} is an analysis and discussion of our 
results.  And we conclude in Section~\ref{sec:conc}.

\section{Related Work}\label{sec:RW}

The problem of low-resource slot and intent detection for languages with limited training data has been the focus of several recent research works. In this section, we discuss some of the most relevant and recent works, including datasets, benchmarks, and models that aim to address this challenge.

\subsection{\textbf{SID Benchmarks and Corpus}}

The table below provides an overview of various datasets used for NLU tasks. These datasets cover a range of languages, domains, intents, and slots, and are widely used to evaluate the performance of NLU models. Some of the prominent datasets include MASSIVE, SLURP, NLU Evaluation Data, ATIS, MultiATIS++, Snips, TOP, MTOP, Cross-lingual Multilingual Task-Oriented Dialog, Microsoft Dialog Challenge, and Fluent Speech Commands. These datasets have been used for tasks such as intent classification, slot filling, and semantic parsing. Overall, these datasets provide a useful resource for researchers to benchmark their models and develop better NLU systems.

\begin{table*}[!ht]
\resizebox{\textwidth}{!}{%
\begin{tabular}{lccccc}
\toprule
\textbf{Name}                                         & \# \textbf{Langs} & \textbf{Utt. per Lang (K)} & \textbf{Domains} & \textbf{Intents} & \textbf{Slots}                                                                                                           \\ \midrule
Airline Travel Information System (ATIS) \citep{price-1990-evaluation} &$1$ &$5.8$ &$1$ &$26$ &$129$ \\
ATIS with Hindi and Turkish \citep{8461905} &$3$ &$1.3$-$5.8$ &$1$ &$26$ &$129$ \\
Cross-lingual Multilingual Task Oriented Dialog \citep{schuster-etal-2019-cross-lingual} &$3$ &$5.08$-$43.3$ &$3$ &$12$ &$11$ \\
Fluent Speech Commands (FSC) \citep{lugosch2019speech} &$1$ &$30$ &- &$31$ &- \\
MASSIVE \cite{https://doi.org/10.48550/arxiv.2204.08582} &$51$ &$19.5$ &$18$ &$60$ &$55$ \\
Microsoft Dialog Challenge \citep{li2018microsoft} &1 &$38.2$ &3 &11 &29 \\
MultiATIS++ \citep{xu-etal-2020-end} &$9$ &$1.4$-$5.8$ &$1$ &$21$-$26$ &$99$-$140$ \\
Multilingual Task-Oriented Semantic Parsing (MTOP) \citep{li-etal-2021-mtop} &$6$ &$15.1$-$22.2$ &$11$ &$104$-$113$ &$72$-$75$ \\
NLU Evaluation Data \citep{liu2019benchmarking} &$1$ &$25,7$ &$18$ &$54$ &$56$ \\
SLURP \citep{bastianelli-etal-2020-slurp} &1 &$16,5$ &18 &60 &55 \\
SNIPS \citep{coucke2018snips} &$1$ &$14.4$ &- &$7$ &$53$ \\
Task Oriented Parsing (TOP) \citep{gupta-etal-2018-semantic-parsing} &$1$ &$44.8$ &$2$ &$25$ &$36$ \\
xSID \cite{van-der-goot-etal-2021-masked} &$13$ &$10$ &$7$ &$16$ &$33$ \\

\bottomrule
\end{tabular}}
\caption{  SID benchmark and datasets with the number of languages covered, number of utterances per language, domain, intent count, and slot count. }\label{table:NLUDatasets}
\end{table*}

\subsection{SID Approaches and Models}

The are many works devoted to the SID tasks. Most of these works are categorized into three approaches: \textit{(1) single model for intent detection, (2)  single model for  slot filling, and (3) joint model}.

\noindent\textbf{(1) Single Model for Intent Detection} refers to developing a single model that can identify the intent behind a user's spoken or written input. This approach involves training a neural network or other machine learning model on a large dataset of labeled examples. Each example consists of user input and its corresponding intent label. The model then uses this training data to learn patterns and features that can accurately predict the intent of new user inputs. For instance,  \newcite{InterspeechRecurrent} proposed 
a recurrent neural network and LSTM models for intent detection in spoken language understanding. In this work, the authors first discuss the limitations of traditional intent detection approaches that rely on handcrafted features and propose using deep learning models to learn features directly from the data.  \newcite{DBLP:journals/corr/abs-2106-04564} investigate the robustness of pre-trained transformers-based models such as BERT and RoBERTa for intent classification in spoken language understanding. They conduct experiments on two datasets, ATIS~\cite{8461905} and SNIPS~\cite{coucke2018snips}, showing that pre-trained transformers perform well on in-scope intent detection.

\noindent \textbf{(2) Single Model for Slot Filling} is an approach that aims to develop a single model capable of identifying slots in spoken language understanding. The model takes a sentence as input and predicts the slot labels for each word in the sentence. Various recurrent neural network (RNN) architectures such as Elman-type \cite{6998838} and Jordan-type \cite{6998838} networks and their variants have been explored to find the most effective architecture for slot filling. Incorporating word embeddings has also been studied and found to improve slot-filling performance significantly.  For example, \newcite{6854368} use LSTM networks with word embeddings for slot filling on the ATIS~\cite{8461905} dataset and achieve state-of-the-art (SOTA) results at the time.
\newcite{goo2018slot} propose a bi-directional LSTM (BLSTM) with an attention mechanism for slot filling on the  ATIS~\cite{8461905} and SNIPS~\cite{coucke2018snips} datasets.

\noindent\textbf{(3) Joint Model} is an approach that aims to jointly model the intent detection and slot-filling tasks in spoken language understanding. This approach trains a single model to predict both the intent and slot labels simultaneously. The model uses the context of the input sentence to predict these labels. Joint models have been shown to achieve SOTA performance on several spoken language understanding datasets. \newcite{xu2013convolutional} propose a joint convolutional neural network (CNN) and RNN model for intent detection and slot filling on the ATIS~\cite{8461905} dataset. They achieved SOTA results at the time. In the same context, \newcite{liu2016attention} proposed an attention-based neural network for joint intent detection and slot filling. The model uses an attention mechanism to weigh the importance of different parts of the input sentence for predicting the intent label and slot labels. \newcite{DBLP:journals/corr/abs-1902-10909} explore the use of the BERT model for joint intent detection and slot filling on ATIS~\cite{8461905} and SNIPS~\cite{coucke2018snips}. They report SOTA results on both datasets.

\section{SID4LR Shared Task}\label{sec:SID4LR}

\noindent{\textbf{Task Formulation.}} Intent detection and slot-filling are critical NLP tasks where, given an utterance, a system is responsible for parsing the user's intent and extracting relevant information to act or reply appropriately. While many neural-based models have achieved SOTA performance for these tasks, their success often depends on large amounts of labeled data. However, many real-world datasets are limited to specific domains and are only available in English or a few other languages. As a result, it is important to reuse existing data in high-resource languages to develop models for low-resource languages, especially since tasks like intent classification and slot-filling require abundant labeled data. 

\noindent{\textbf{Shared Task Problem Statement.} This shared task of SID aims to address the challenges of performing SID for low-resource language varieties for the following languages: Swiss German, South Tyrolean, and Neapolitan. 
The training data provided consists of the Cross-lingual Slot and Intent Detection (xSID\textsubscript{0.4}) corpus \cite{van-der-goot-etal-2021-masked}, a cross-lingual spoken language understanding dataset, covering $12$ languages (Arabic, Chinese, Dutch, Danish, English, German, Indonesian, Italian, Japanese, Kazakh, Serbian, Turkish) from six language families with English training. The task allowed the use of pre-trained models and external data including data from the target language.}

\noindent{\bf{Evaluation Metric.}} The primary evaluation metric for slot filling is the span F\textsubscript{1} score, where both span and label must match exactly, and accuracy is used to evaluate intent detection where it is calculated through the ratio of the number of correct predictions of intent to the total number of sentences. More details regarding the shared task can be found in  \citet{2023-findings-vardial}.

\section{Data}\label{sec:data}

\noindent\textbf{Shared Task Data.} The xSID\textsubscript{0.4} \cite{van-der-goot-etal-2021-masked} corpus comprises cross-lingual SLU evaluation datasets covering $13$ languages from six language families. The training dataset contains $43,605$ sentences, the development set contains $300$ sentences, and the test set contains $500$ sentences. The corpus contains sentences from Snips and Facebook, which were translated into all $13$ target languages, resulting in a cross-lingual SLU evaluation dataset. All examples are annotated with their intent and corresponding slots. Tables~\ref{tab:data1} and ~\ref{tab:dat2} provides examples of annotations with intent and slots. We converted the dataset into a JSON format that includes intents and slots. A sample of the resulting JSON format is show below.

\begin{table}
\centering
\resizebox{\columnwidth}{!}{%
\begin{tabular}{lrrr}
\toprule
\textbf{Language} &  \# \textbf{Train} &  \# \textbf{Valid} &  \# \textbf{Test} \\
\midrule
   ar &  $42,157$ &    $300$ &   $500$ \\
   da &  $43,605$ &    $300$ &   $500$ \\
   de &  $43,605$ &    $300$ &   $500$ \\
   en &  $43,605$ &    $300$ &   $500$ \\

   id &  $42,157$ &    $300$ &   $500$ \\
   it &  $43,605$ &    $300$ &   $500$ \\
   ja &  $29,073$ &    $150$ &   $250$ \\
   kk &  $42,157$ &    $300$ &   $500$ \\
   nl &  $43,605$ &    $300$ &   $500$ \\
   sr &  $43,605$ &    $300$ &   $500$ \\
   tr &  $43,605$ &    $300$ &   $500$ \\
   zh &  $42,157$ &    $300$ &   $500$ \\
\bottomrule
\end{tabular}}
\caption{ Number of samples in the train, validation, and test sets for each language in the  multilingual dataset xSID\textsubscript{0.4}, where the language codes are represented by two-letter ISO codes. The dataset includes 12 languages: Arabic (ar), Danish (da), German (de), English (en), Indonesian (id), Italian (it), Japanese (ja), Kazakh (kk), Dutch (nl), Serbian (sr), Turkish (tr), and Chinese (zh).}\label{tab:bAbI}
\end{table}

\begin{table}[!htp]\centering
\caption{Example of data summary}\label{tab: }
\scriptsize
\resizebox{1\columnwidth}{!}{%
\begin{tabular}{lrr}\toprule
\# \textcolor{blue}{\texttt{text}}: & \textbf{show all reminders} \\[0.2cm]
\# \textcolor{blue}{\texttt{intent}}: &\texttt{reminder/show\_reminders} \\[0.2cm]
\# \textcolor{blue}{\texttt{slots}}: & \texttt{5:8:reminder/reference, 9:18:reminder/noun} \\
\bottomrule
\end{tabular}}
\end{table}\label{tab:data1}

\begin{table}[!htp]\centering
\caption{In-depth data example}\label{tab: }
\scriptsize
\resizebox{1\columnwidth}{!}{%
\begin{tabular}{lrrr}\toprule
1 \textcolor{red}{show} &reminder/show\_reminders &O \\[0.1cm]
2 \textcolor{red}{all} &reminder/show\_reminders &B-reference \\[0.1cm]
3 \textcolor{red}{reminders} &reminder/show\_reminders &O \\
\bottomrule
\end{tabular}}
\end{table}\label{tab:dat2}

\begin{lstlisting}[language=json,firstnumber=1,caption=Example of preprocessed data,captionpos=b]
{
    'text':'show all reminders',
    'slots': 'reference:all',
    'intent': 'reminder/show_reminders',
    '__index_level_0__': 0
 }
\end{lstlisting}
\label{data:json}

\noindent\textbf{External Data.} As mentioned, Swiss German, South Tyrolean, and Neapolitan are low-resource languages with limited available labeled data. To address this challenge, we incorporate unlabeled data from different sources to augment our training data. We describe these external sources next.

\noindent\textbf{SwissCrawl} \cite{linder2020crawler}, a corpus of over $500,000$ Swiss German sentences gathered from web crawling between September and November $2019$. The sentences are representative of how native speakers write in forums and social media and may contain slang and ascii emojis.

\noindent \textbf{DiDi Corpus} \cite{inproceedings} is a multilingual language corpus of $600,000$ tokens from Facebook users in South Tyrol, Italy. It includes CMC texts, socio-demographic data, and linguistic annotations on thread, text, and token level. The corpus is mainly German and Italian, with English also present, and has been manually anonymized and annotated.

\noindent\textbf{OSCAR Corpus} \cite{caswell-etal-2021-quality} is a large multilingual corpus created by scraping the web and includes texts in more than $200$ languages. The OSCAR Corpus includes texts in Neapolitan, which is a Romance language spoken in the southern part of Italy, particularly in the region of Campania. The Neapolitan texts in the corpus consist of around $4.4$ million tokens, making it one of the largest resources available for this language.

\section{Pre-trained Language Models}\label{sec:models}
In this study, we evaluate several popular multilingual Transformer-based language models, including mBERT, XLM-R, SBERT, LaBSE, LASER, and mT0. These models are capable of effectively capturing cross-lingual embeddings, enabling transfer learning across multiple languages. Below we provide a description of each model used in our experiments on the training dataset. 

   \noindent\textit{\textbf{mBERT.}} is the multilingual version of BERT~\cite{devlin2019bert}, which is an encoder model with bidirectional representations from Transformers trained with a denoising objective.  mBERT is trained on Wikipedia for $104$ languages including German and Italian.

 \noindent\textit{\textbf{XLM-R.}}~\cite{conneau-etal-2020-unsupervised} is a transformer-based multilingual masked language model pre-trained on more than $2$TB  of filtered CommonCrawl data in $100$ languages, including languages including German and Italian. XLM-R uses a Transformer model
~\cite{vaswani2017attention} trained with a multilingual masked language model XLM \cite{lample2019cross}.

\noindent{\textbf{sBERT.}} Sentence-BERT (SBERT)~\cite{reimers2019sentence}, is a modification of the pretrained BERT~\cite{devlin2019bert}  model that uses siamese and triplet network structures to derive semantically meaningful sentence embeddings that can be compared using cosine-similarity. As we work under a multilingual context, we use the  multilingual versions from previously monolingual SBERT models~\cite{reimers2020making} which is trained for sentence embedding in $50$+ languages from various language families.

\noindent{\textbf{LaBSE.}} Language-agnostic BERT Sentence Encoder (LaBSE)~\cite{feng2020languageagnostic} is a BERT-based model  trained to generate sentence embeddings in $109$ different languages.  The model's pre-training approach involves a combination of masked language modeling and translation language modeling. The pre-training process combines masked language modeling with translation language modeling. LaBSE is useful for producing sentence embeddings in multiple languages and performing bi-text retrieval.

\noindent{\textbf{LASER.}} Language-Agnostic Sentence Representations (LASER)~\cite{feng2020language} is a contextualized language model based on a BiLSTM encoder trained on parallel data from OPUS website~\cite{tiedemann2012parallel} using a translation objective. The LASER model can handle $200$ different languages.

\noindent{\textbf{mT0.}} \cite{muennighoff2022crosslingual} is a group of sequence-to-sequence models that come with different sizes from $300$M to $13$B  parameters trained to investigate the cross-lingual generalization through multitask finetuning. mT0  can execute human instructions in many languages without any prior training. The models are fine-tuned from pre-existing mT5 ~\cite{xue2020mt5} multilingual language models using a cross-lingual task mixture called xP3. These refined models are capable of cross-lingual generalization to unseen languages.

\section{Experiments and Settings}\label{sec:exp}

\noindent{\textbf{Training on English Data.}} As a baseline setting, we train all the pre-trained models described in Section~\ref{sec:models} on the English part of the multilingual dataset  xSID\textsubscript{$0.4$}~\cite{van-der-goot-etal-2021-masked}  and evaluate them on Swiss German, South Tyrolean, and Neapolitan under a zero-shot setting.


\noindent{\textbf{Training on German/Italian Data.}} Our second approach aims to train all the pre-trained models on the language family of low-resource languages (i.e., German for Swiss German and South Tyrolean, and Italian for Neapolitan, respectively) under the zero-shot setting. So, we extract the German and Italian SID data from xSID\textsubscript{$0.4$}, and then fine-tune all our models on both datasets. Then, we evaluate the German models on GSW and ST tasks and the Italian models on the NAP task.



\noindent{\textbf{Training on Multilingual Dataset.}} Next, we explore a third training approach that involves the full multilingual xSID\textsubscript{$0.4$} dataset. To do so, we combine all the $12$ available languages in the xSID\textsubscript{$0.4$} dataset and fine-tune our pre-trained models on this combined dataset. We then evaluate each target using a zero-shot setting. This approach allows us to train on larger and more diverse datasets. In total, we generate $502,936$ training sentences across all languages in the dataset. 

\noindent{\textbf{Paraphrase and Machine Translation.}} To improve the performance of our pretrained models, we also explore the impact of data augmentation techniques such as paraphrasing and machine translation. Specifically, we aime to examine how these techniques can enhance the performance of our models on cross-lingual SLU tasks. To this end, we experiment with different data augmentation strategies, including paraphrasing and machine translation. Paraphrasing is performed using the quality-guided controlled paraphrase generation (QCPG) model~\cite{bandel-etal-2022-quality}, resulting in a total of $130,815$ sentences in English. These sentences are then translated into German and Italian using the OPUS-MT model \cite{TiedemannThottingal:EAMT2020}, creating cross-lingual datasets for our experiment. 

To further augment our training data for low-resource languages, we leverage Meta AI's No Language Left Behind (NLLB), which provides open-source models capable of high-quality translations between $200$ languages (including low-resource languages \cite{nllb2022}). To create our new training data using the NLLB model, we first use FastText to detect the language codes of our target languages. Next, we utilize NLLB models to translate the English training data into the predicted language codes. The language codes identified for our target languages are \textit{deu{\_}latn} for Swiss German, \textit{est{\_}latn} for South Tyrolean, and \textit{ita{\_}Latn} for Neapolitan. We generate $43,605$ sentences for each of the three languages. It is worth noting that we ensure that the labels for each sentence remain the same throughout the paraphrasing and machine translation process to maintain the integrity of the data.

\noindent{\textbf{Training on External Data.}
Since the language models we employ do not have a strong representation of the low-resource languages used on the task, we leverage large corpora of each of the low-resource languages into the training process. By incorporating external datasets, the models are exposed to more comprehensive information about the semantics of each low-resource language, enabling them to better capture the nuances and complexities of the target languages.

\noindent{\textbf{Training MT0}}
As discussed in Section \ref{sec:models}, the MT0 models share the same architecture as MT5/T5 models, i.e., they are encoder-decoder models. Therefore, we train them for intent classification and slot detection using the data preprocessing approach described in Section \ref{sec:data}. We utilize the PEFT library provided by Huggingface \cite{peft} to train the mT0-small, mT0-Base, and mt0-Large models. Our approach involves using LORA \cite{DBLP:journals/corr/abs-2106-09685}, which allows us to achieve SOTA performance while consuming significantly less memory. For the mT0-xxl models, we utilize DeepSpeed \cite{10.1145/3394486.3406703} with CPU offloading to train a model with 13B parameters on a 40GB A100 GPU. 

\noindent{\textbf{Combining Models.}}} 
In recent studies, joint learning techniques that combine multiple classification approaches have produced promising results \cite{bilat2020cross}. These approaches involve concatenating the outputs of individual models and passing the resulting output through multiple neural network layers, allowing the resulting network to be trained jointly. In this part of our experiments, we investigate the effectiveness of this approach in zero-shot settings by combining multilingual models. Specifically, we combine LASER embeddings, from the LASER model, with other multilingual models including mBERT, sBERT, LaBSE and XLMR.
\section{Results and Discussions}\label{sec:results}
\begin{table*}[t]\centering
\resizebox{\textwidth}{!}{%
\begin{tabular}{llccccccccc}\toprule
\textbf{Setting} &\textbf{Lang.} &\textbf{mBert} &\textbf{LS} &\textbf{LL} &\textbf{LX} &\textbf{mT0-small} &\textbf{mT0-base} &\textbf{mT0-large} &\textbf{mT0-xxl} \\\midrule

\multirow{3}{*}{\textbf{English}} &GSW &$51.67$ &$45.30$ &$52.70$ &$48.30$ &$69.20$ &$70.20$ &$69.00$ &\underline{80.00} \\
&ST &$61.00$ &$58.00$ &$66.70$ &$61.70$ &$74.50$ &$76.20$ &$79.10$ &\underline{89.00} \\
&NAP &$61.00$ &$55.30$ &$56.00$ &$67.0$ &$71.30$ &$72.00$ &$75.00$ &\underline{76.33} \\\hline

\multirow{2}{*}{\textbf{German}} &GSW &$59.00$ &$74.00$ &$68.00$ &$80.70$ &$69.30$ &$73.33$ &$80.33$ &\underline{84.33} \\
&ST &$59.70$ &$55.70$ &$59.00$ &$51.00$ &$83.33$ &$88.33$ &$84.66$ & \underline{92.00} \\\hline
\textbf{Italian} &NAP &$65.30$ &$63.30$ &$63.70$ &$55.70$ &$77.66$ &$84.66$ &$83.33$ & \underline{86.00} \\\hline

\multirow{3}{*}{\textbf{Multilingual}} &GSW &$59.70$ &$62.70$ &$59.70$ &$53.30$ &$75.00$ &$76.33$ &$84.00$ &\underline{$89.00$} \\
&ST &$60.70$ &$54.70$ &$58.00$ &$56.30$ &$88.33$ &$85.66$ &$90.66$ &\underline{$94.00$} \\
&NAP &$61.30$ &$55.70$ &$59.00$ &$60.30$ &$82.66$ &$84.66$ &$86.00$ &\underline{$87.00$}  \\\hline

\multirow{3}{*}{\textbf{Paraphrase+MT}} &GSW &$45.30$ &$37.30$ &$58.00$ &$64.00$ &$79.00$ &$83.00$ &$84.33$ &\textbf{91.00} \\
&ST &$61.70$ &$61.30$ &$60.70$ &$60.70$ &$90.66$ &$93.00$ &$90.00$ &\textbf{95.66} \\
&NAP &$63.70$ &$60.00$ &$58.70$ &$60.00$ &$85.66$ &\textbf{89.00} &$87.33$ &$88.33$ \\
\bottomrule
\end{tabular}}
\caption{Accuracy results for intent classification on the validation set.  \textbf{Baseline:} mBERT~\cite{devlin2019bert}. \textbf{LS:} LASER~\cite{feng2020language}+sBERT~\cite{reimers2019sentence}.  \textbf{LL:} LASER+LaBSE~\cite{feng2020languageagnostic}. \textbf{LX:} LASER+XLM-R~\cite{lample2019cross}. \underline{\textbf{Underline}:} Best-performing models for each setting. \textbf{Bold:}~Best F\textsubscript{1} score across all the experiments and settings. }  \label{tab:dev_res}
\end{table*}

\noindent{\textbf{Evaluation on Validation Data.}} We present the accuracy scores of all our models across various settings. Table~\ref{tab:dev_res} presents the evaluation results for 
the intent classification task on the validation set. Our transformer-based models, with different experimental settings, outperform the baseline on all the target languages. For instance, \textbf{mT0-base} outperforms the baseline (mBERT) with an average of $+16.49$, $+22.93$, $+17.90$ for GSW, ST, and NAP, respectively. Notably, our best combination was the mT0-xxl model under the multilingual setting. It achieves the best results of $89.00$, $94.00$, and $87.00$, improving the baseline with $+29.30$, $+33.30$, and $+25.70$ Accuracy point in the three target languages. 

The  results  of the slot filling task on the validation set are shown in Table~\ref{tab: slots}. Our transformer-based models perform better than the baseline across all target languages when tested under different experimental settings. Our best-performing model, mT0-large, achieves the most outstanding results using the Multilingual settings with F\textsubscript{1} scores of $60.30$, $55.00$, and $52.30$ in the three target languages. These results represent a notable improvement over the baseline, with an increase of $+30.88$, $+4.65$, and $+0.90$ F\textsubscript{1} points in the three target languages. 

Our results on the validation data suggest that larger models generally achieve better performance, implying that higher parameter counts result in better cross-lingual and zero-shot setting performance. Moreover, as the mT0 models are fine-tuned from pre-existing mT5 multilingual language models, they are capable of performing cross-lingual generalization on unseen languages. This capability may be a possible reason for the mT0 models outperforming other models in zero-shot settings.

\begin{table}[t]\centering

\centering
 \renewcommand{\arraystretch}{1.1}
\resizebox{1\columnwidth}{!}{%
\begin{tabular}{llrrrrr}\toprule
\textbf{Setting} &\textbf{Lang.} &\textbf{Baseline} &\textbf{mt0-small} &\textbf{mt0-base} &\textbf{mt0-large} \\\midrule
\multirow{3}{*}{English} &GSW &26.23  &25.42  &34.00  &\textbf{40.32 } \\
&ST &44.61  &32.40  &44.00  &\textbf{54.30 } \\
&NAP &48.01  &42.20  &47.90  &\textbf{49.00 } \\\hline
\multirow{3}{*}{Multi-langl} &GSW &29.42  &28.90  &42.30  &\textbf{60.30 } \\
&ST &50.35  &43.40  &53.40  &\textbf{55.00 } \\
&NAP &51.40  &49.00  &50.30  &\textbf{52.30 } \\
\bottomrule

\end{tabular}}

\caption{Slot-f1 results for Slot Filling on the validation set. Bold entries are the best-performing models for each experiment and setting.}\label{tab: slots}
\end{table}
\noindent{\textbf{Official Shared Task (Test) Results.}} Our findings regarding the performance of larger models are also observed in the test set. Table 6 presents the evaluation results for both slot filling  and intent classification tasks across all three target languages. Our mT0 models strongly outperform the baseline models. Specifically, our mT0 models outperformed the baseline models in all target languages for the intent classification task, highlighting the effectiveness of larger models for intent classification. Moreover, our mT0 models also outperform the baseline models in two of the target languages for slot filling task, further indicating the superiority of larger models for sentence-level classification tasks. The improvement in scores for intent classification is more evident than for slot filling. The larger improvement in scores for intent classification may be correlated with the fact that for our data augmentation experiment on paraphrasing and machine translation, we were only able to augment data for intent classification, resulting in a larger improvement in performance for this task compared to slot filling.

It is worth noting that we use the validation set for model selection, which resulted in higher scores than those achieved on the test set. This is because the validation data is similar to the data used during training, while the test data is entirely new and unseen. As a result, the test scores may be lower due to differences in the distribution of data between the training and test sets. Nevertheless, our mT0 models consistently outperform the baseline models on the test set, providing further evidence for the effectiveness of larger models in SID tasks.

\begin{table}

\centering
 \renewcommand{\arraystretch}{1.1}
\resizebox{\columnwidth}{!}{%
\begin{tabular}{llcc}
\hline
\textbf{Task} & \textbf{Lang.} & \textbf{\begin{tabular}[c]{@{}l@{}}Baseline\end{tabular}} & \textbf{\begin{tabular}[c]{@{}l@{}}mT0-large\end{tabular}} \\ \hline
                     & ST  & $44.61$ & $\textbf{46.41}$ \\
\textbf{Slots}    & GSW & $26.23$ & $\textbf{27.39}$ \\
                     & NAP & $\textbf{48.01}$ & $38.82$ \\ \hline
                     & ST  & $61.00$ & $\textbf{89.40}$ \\
\textbf{Intents} & GSW & $51.67$ & $\textbf{81.60}$ \\
                     & NAP & $61.00$ & $\textbf{85.40}$ \\ \hline
\end{tabular}}
\caption{Results on the test set for both SID tasks. \textbf{Bold} entries indicate the model's performance compared to the baseline model.} \label{tab:test_res}
\end{table}
\section{Conclusion}\label{sec:conc}

We described our contribution to the SID4LR~\cite{2023-findings-vardial} shared tasks. Our models target both the slot and intent sub-task in three proposed low-resource languages,  namely, Swiss German, South Tyrolean, and Neapolitan. We test the utility
of existing pretrained language models such as mT0~\cite{muennighoff2022crosslingual}  on the intent detection and slot filling tasks. We show that such models can lead to improving the results of the baseline with an average of +$27$ F\textsubscript{1} points. In the future, we intend to use mT0 to jointly model the intent detection and slot filling tasks  for improving overall performance.

\section*{Acknowledgements}
We gratefully acknowledge support from the Natural Sciences and Engineering Research Council of Canada (NSERC; RGPIN-2018-04267), the Social Sciences and Humanities Research Council of Canada (SSHRC; 435-2018-0576; 895-2020-1004; 895-2021-1008), Canadian Foundation for Innovation (CFI; 37771), Compute Canada (CC),\footnote{\href{https://www.computecanada.ca}{https://www.computecanada.ca}} UBC ARC-Sockeye,\footnote{\href{https://arc.ubc.ca/ubc-arc-sockeye}{https://arc.ubc.ca/ubc-arc-sockeye}} and Advanced Micro Devices, Inc. (AMD). Any opinions, conclusions or recommendations expressed in this material are those of the author(s) and do not necessarily reflect the views of NSERC, SSHRC, CFI, CC, AMD, or UBC ARC-Sockeye.

\bibliography{anthology,custom}
\bibliographystyle{acl_natbib}



\end{document}